\documentclass{isprs} 
\usepackage{amsmath}
\usepackage{subfigure}
\usepackage{setspace}
\usepackage{geometry} 
\usepackage{epstopdf}
\usepackage{comment}

\usepackage[labelsep=period]{caption}  
\usepackage{booktabs}
\usepackage{multirow}
\usepackage[british]{babel} 
\usepackage[hang]{footmisc}
\usepackage[dvipsnames]{xcolor}

\usepackage{xspace}
\makeatletter
\DeclareRobustCommand\onedot{\futurelet\@let@token\@onedot}
\def\@onedot{\ifx\@let@token.\else.\null\fi\xspace}

\makeatother

\usepackage{amsmath}

\begin{document}

\title{RAMBA: 4D Radar Mapping by Bundle Adjustment}

\author{Jianzhu Huai\textsuperscript{1}, Yiwen Chen\textsuperscript{1}, Binliang Wang\textsuperscript{1}}%
\address{
	\textsuperscript{1 }State Key Lab of Info Engineering in Surveying, Mapping and Remote Sensing,\\
    129 Luoyu Road, Hongshan District, Wuhan, Hubei, China \\
jianzhu.huai@whu.edu.cn; chenyiwen@whu.edu.cn; blwang@whu.edu.cn
}


\abstract{
4D radar is increasingly attractive for robotic mapping because it provides range, azimuth, elevation, and Doppler measurements while remaining robust in adverse visibility conditions. 
Although recent radar and radar--inertial odometry methods have achieved promising online state estimation performance, offline global map refinement for 4D radar remains underexplored. 
This paper presents \textbf{RAMBA}, a radar bundle-adjustment framework for globally consistent 4D radar mapping. 
Given initial poses and radar frames from a radar--inertial odometry front-end, RAMBA jointly refines radar frame states using covariance-weighted geometric residuals, IMU preintegration factors, and radar ego-velocity constraints. 
The geometric residuals extend pairwise GICP to a multi-frame optimization by forming voxel-based correspondences across selected frames and weighting each residual with point covariances. 
To improve robustness against drift and revisits, RAMBA enforces temporal consistency during correspondence formation while explicitly supporting loop-closure constraints.
Experiments on the ColoRadar and SNAIL Radar datasets show that RAMBA improves map consistency and usually enhances trajectory accuracy over radar--inertial odometry and pose-graph optimization baselines.
}

\maketitle

\section{Introduction}\label{sec:intro}

\sloppy

4D radars have attracted increasing interest for robotic perception because they remain effective in adverse conditions such as darkness, dust, smoke, rain, and fog. Compared with conventional automotive radars that mainly provide planar coordinates and relative Doppler velocity, modern 4D radars also sense elevation, which makes them more suitable for geometric odometry and mapping.

Recent studies have shown strong progress in radar odometry and radar--inertial state estimation \cite{doerRadarVisualInertial2021,zhuang4DIRIOM4D2023}. Learning-based radar mapping has also been investigated \cite{mopideviRMapMillimeterWaveRadar2023}. However, most existing radar methods focus on online pose estimation or learned scene reconstruction. In contrast, offline map refinement for radar remains largely unexplored, even though analogous LiDAR problems have benefited substantially from global optimization methods such as \cite{wiesmannEfficientLiDARBundle2024,liEfficientDistributedLargeScale2025}.

In this paper, we propose \textbf{RAMBA}, an offline 4D radar mapping framework based on bundle adjustment. Starting from initial poses and radar frames produced by a radar--inertial odometry front-end, we refine the radar frame states to improve global mapping consistency, measured by covariance-weighted point-to-point distances. In essence, our method extends pairwise generalized iterative closest point (GICP) to the multi-frame setting. Candidate correspondences are formed within voxels of a voxel grid built from all selected frames, and each residual is weighted by the sum of the two point covariances. The geometric constraints are jointly optimized with IMU preintegration and radar ego-velocity constraints.

To reduce false associations caused by drift and revisits, RAMBA enforces temporal consistency when forming correspondences and explicitly allows constraints around loop closures. We evaluate the method on the ColoRadar and SNAIL Radar datasets. The proposed refinement consistently improves map quality and usually improves trajectory accuracy over the initial radar--inertial odometry and pose graph optimization. To the best of our knowledge, this is the first geometric offline bundle-adjustment framework for consistent 4D radar mapping.

\subsection{Related Work}

Radar odometry and radar--inertial estimation have advanced rapidly in recent years \cite{doerRadarVisualInertial2021,zhang4DRadarSLAM4DImaging2023,michalczykTightlycoupledEKFbasedRadarinertial2022,zhuang4DIRIOM4D2023}. These methods provide promisingly accurate pose estimates in real time, but the resulting maps can still suffer from accumulated drift and revisit inconsistency.
For spinning radars, DR-BA \cite{lisus2026drba} refines mapping consistency by making all overlapping radar intensities agree in the same map frame.
For single-chip 4D radars, learning-based radar mapping method \cite{mopideviRMapMillimeterWaveRadar2023} addresses radar map densification but uses the ground truth trajectory.

In the LiDAR domain, offline trajectory and map refinement have been studied extensively. Representative examples include graph-based and bundle-adjustment-based methods \cite{liGraphOptimalityAwareStochastic2025,wiesmannEfficientLiDARBundle2024,koideGloballyConsistentTightly2022,liEfficientDistributedLargeScale2025}. These methods demonstrate the value of jointly refining multiple frames using geometric constraints. However, transferring them directly to radar is not straightforward, because radar points are substantially noisier and often do not support reliable normal estimation. As a result, surfel constraints such as those in \cite{liEfficientDistributedLargeScale2025} and point-to-plane constraints such as those in \cite{wiesmannEfficientLiDARBundle2024} are less suitable for radar mapping.

Our method is most closely related in spirit to LiDAR bundle adjustment \cite{wiesmannEfficientLiDARBundle2024}, but differs in several important aspects. First, we use covariance-weighted point-to-point distances like \cite{koideGLIM3DRangeinertial2024} rather than point-to-plane distances, which are better suited to the noisy range measurements of radar frames. Second, we construct constraints from all valid point pairs collected in a world-frame voxel grid, rather than matching each keyframe only to a local submap, since radar point clouds are typically much sparser than LiDAR point clouds. Third, we operate only on stationary radar points filtered by the front-end GNC process, which helps suppress dynamic clutter by exploiting the Doppler information naturally available in radar measurements.

\subsection{Overview of the Proposed Method}

The inputs to RAMBA are the initial states and undistorted radar frames from a radar--inertial odometry front-end, e.g., 4D iRIOM \cite{zhuang4DIRIOM4D2023}. We first select keyframes according to overlap and detect loop candidates using PointNetVLAD \cite{chen4DRadarPRContextaware2025}. Verified loop closures are then used to guide offline refinement.

For each keyframe, we first estimate point covariances using a voxel grid built from its temporal neighborhood. For computational simplicity, only the diagonal entries of the covariance matrices are retained. We then construct point-to-point constraints for mapping using the current keyframe states and a world-frame voxel grid. Finally, we optimize the keyframe poses, world-frame velocities, and IMU biases by jointly minimizing the voxel-based generalized point-to-point cost, IMU preintegration residuals, and ego-velocity residuals. After keyframe optimization, we recover all frame poses by pose graph optimization using the optimized keyframe constraints together with adjacent-frame relative pose constraints.

\section{Method}\label{sec:method}

\subsection{Keyframe Selection and Loop Verification}\label{subsec:keyframe}

Before offline optimization, we run a radar--inertial odometry front-end and retain only stationary radar points using the GNC-based filtering already available in the front-end. We then select keyframes based on overlap. A sliding-window map is built from the most recent 21 keyframes and represented by a voxel grid. A new frame is promoted to a keyframe when its overlap with the current sliding-window map falls below a threshold. In our experiments, we use a threshold of 0.6 for SNAIL Radar and 0.5 for ColoRadar.

For each keyframe, we aggregate its 21-frame temporal neighborhood and build a local voxel grid for covariance estimation. Each point is assigned a covariance $\mathbf{C}_{f_{ij}}$ computed from the points in its corresponding voxel. To simplify computation, we retain only the diagonal entries and denote the resulting diagonal covariance again by $\mathbf{C}_{f_{ij}}$. The corresponding world-frame covariance is
\begin{equation}
    \mathbf{C}_{ij} = \mathbf{R}_{WB,j}\mathbf{C}_{f_{ij}}\mathbf{R}_{WB,j}^\top .
\end{equation}

Loop candidates are detected by PointNetVLAD \cite{chen4DRadarPRContextaware2025} and verified by a local GICP alignment. Specifically, we build two 5-frame submaps centered at the query frame and the candidate loop frame, respectively. Each point is assigned the covariance of its voxel in the corresponding submap, and the relative pose is refined using covariance-weighted point-to-point constraints. Only verified loop closures are retained.

The voxel size used in the associated voxel-grid operations is 0.12\,m on ColoRadar and 0.5\,m on SNAIL Radar.

\subsection{Voxel-Based Feature Association}\label{subsec:feature}

Let the optimized state of keyframe $j$ be
\begin{equation}
    \mathbf{x}_j =
    \left(
    \mathbf{R}_{WB,j},
    \mathbf{t}_{WB,j},
    \mathbf{v}_j^W,
    \mathbf{b}_{g,j},
    \mathbf{b}_{a,j}
    \right),
\end{equation}
where $\mathbf{R}_{WB,j}$ and $\mathbf{t}_{WB,j}$ denote the pose of the IMU/body frame in the world frame, $\mathbf{v}_j^W$ is the body velocity in the world frame, and $\mathbf{b}_{g,j}$ and $\mathbf{b}_{a,j}$ are the gyroscope and accelerometer biases.

Given the current keyframe states, we transform all keyframe points into the world frame and insert them into a voxel grid. To avoid inconsistency in optimization, each voxel stores at most one point from each frame.
Suppose voxel $V_i$ contains points contributed by several keyframes. Let $\mathbf{p}_{f_{ij}}$ denote the radar-frame point from frame $j$ that falls into voxel $V_i$. Its world-frame position is
\begin{equation}
    \mathbf{p}_{ij} = \mathbf{R}_{WB,j}(\mathbf{R}_{BR} \mathbf{p}_{f_{ij}} + \mathbf{p}_{BR}) + \mathbf{t}_{WB,j},
\end{equation}
where $\mathbf{R}_{BR}$ and $\mathbf{p}_{BR}$ are the radar extrinsics with respect to the body frame.

From each voxel, only point pairs whose frame pair is considered valid are used to form the covariance-weighted point-to-point constraints. Specifically, a frame pair is regarded as valid if the two frames have sufficient spatial overlap, measured by a voxel-overlap ratio greater than 0.1, and additionally satisfy one of the following two conditions: 1) their timestamps differ by less than 30\,s; or 2) they are temporally close to a verified loop-frame pair, with both frame timestamps differing from the loop pair by less than 30\,s. The spatial-overlap condition helps ensure stable registration, the temporal-neighborhood condition suppresses false matches at revisits caused by odometry drift, and the loop-related condition preserves constraints induced by loop closure.

Let $\mathcal{P}_i$ denote the set of valid frame pairs associated with voxel $V_i$. The geometric cost contributed by voxel $V_i$ is
\begin{equation}
    c_i(\{\mathbf{x}_j\})
    =
    \sum_{(j, k)\in\mathcal{P}_i}
    \left\|
    \mathbf{p}_{ij} - \mathbf{p}_{ik}
    \right\|^2_{(\mathbf{C}_{ij}+\mathbf{C}_{ik})^{-1}} .
\end{equation}
This cost extends pairwise GICP to the multi-frame setting by enforcing consistency among points from different frames that fall into the same voxel.

\subsection{Bundle-Adjustment Objective}\label{subsec:objective}
The full objective combines the voxel-based geometric cost with IMU preintegration residuals, ego-velocity residuals, and priors on the first keyframe pose and biases:
\begin{equation}
\begin{aligned}
    J
    &=
    \sum_i c_i
    + \sum_{j=2}^{M}
    \mathbf{r}_{\mathrm{imu},j}^\top
    \mathbf{\Sigma}_{\mathrm{imu},j}^{-1}
    \mathbf{r}_{\mathrm{imu},j} \\
    &\quad
    + \sum_{j=1}^{M}
    \mathbf{r}_{\mathrm{ego},j}^\top
    \mathbf{\Sigma}_{\mathrm{ego},j}^{-1}
    \mathbf{r}_{\mathrm{ego},j} \\
    &\quad
    + \mathbf{r}_{\mathrm{pose},1}^\top
    \mathbf{\Sigma}_{\mathrm{pose},1}^{-1}
    \mathbf{r}_{\mathrm{pose},1}
    + \mathbf{r}_{\mathrm{bias},1}^\top
    \mathbf{\Sigma}_{\mathrm{bias},1}^{-1}
    \mathbf{r}_{\mathrm{bias},1},
\end{aligned}
\end{equation}
where $M$ is the number of keyframes.
The IMU factor $\mathbf{r}_{\mathrm{imu},j}$ connects adjacent keyframe states and constrains pose, velocity, and bias consistency through preintegration.

The ego-velocity residual on keyframe $j$ is
\begin{equation}
    \mathbf{r}_{\mathrm{ego},j}
    =
    \mathbf{R}_{BR}^\top
    \left(
    \mathbf{R}_{WB,j}^\top \mathbf{v}_j^W
    + \boldsymbol{\omega}_j^B \times \mathbf{p}_{BR}
    \right)
    -
    \widetilde{\mathbf{v}}_j^R ,
\end{equation}
where $\boldsymbol{\omega}_j^B$ is the body angular velocity, and $\widetilde{\mathbf{v}}_j^R$ is the measured radar ego velocity. Its covariance is determined using the GNC-based weighting strategy of \cite{zhuang4DIRIOM4D2023}.

Finally, $\mathbf{r}_{\mathrm{pose},1}$ fixes the gauge freedom by constraining the first keyframe pose, and $\mathbf{r}_{\mathrm{bias},1}$ imposes a prior on the first keyframe biases.

\subsection{Optimization and Recovery of Non-Keyframes}\label{subsec:opt}
We minimize the objective using a nonlinear least-squares solver with automatic differentiation. The optimization alternates between two steps:
\begin{enumerate}
    \item update voxel-based correspondences using the current keyframe states;
    \item refine the keyframe states using the geometric, IMU, and ego-velocity factors.
\end{enumerate}
In the current implementation, we run six inner iterations and six outer iterations.

After optimizing the keyframes, we recover the poses of all non-keyframes by solving a pose graph optimization problem over all frames. The optimized keyframe poses are used as constraints, and adjacent-frame relative pose constraints from the front-end are added between consecutive frames. This produces a refined trajectory for all radar frames and is used to build the final point-cloud map.

\section{Experiments}\label{sec:experiments}
\subsection{Experimental Setup}\label{subsec:setup}

We evaluate RAMBA on the ColoRadar dataset's single-chip radar \cite{kramerColoRadar2022} and the SNAIL Radar dataset's ARS548 radar \cite{huaiSNAILRadarLargescale2025}, which together represent short-range and long-range 4D radar sensing. The initial states are provided by our implementation of the IRIOM front-end \cite{zhuang4DIRIOM4D2023}.

We report both trajectory and map-quality metrics. For trajectory accuracy, we use ATE RMSE and rotation RPE RMSE over traveled distances of 40, 60, 80, 100, and 120\,m. For map quality, we use the Chamfer-L1 distance (CD-L1) between the point cloud aggregated with estimated poses and that aggregated with ground-truth poses. The CD-L1 implementation follows \cite{panPINSLAMLiDARSLAM2024}.

Before evaluating full-sequence optimization, we first conduct a unit test on SNAIL Radar sequences by randomly perturbing the pose of a query frame while keeping the reference frame fixed at ground truth. Both the query and reference points use covariances computed from a 21-frame temporal neighborhood. Optimizing only the query pose with the proposed covariance-weighted point-to-point objective almost always recovers the reference pose within 0.1\,m and 0.2$^\circ$, which validates the effectiveness of the geometric cost for refining radar relative poses.

\subsection{Results on ColoRadar}\label{subsec:coloradar}

On ColoRadar \cite{kramerColoRadar2022}, we evaluate the \textit{edgar\_classroom\_run0} and \textit{outdoors\_run0} sequences. The place-recognition module finds one verified revisit in each sequence.

\begin{figure}[!htbp]
    \centering
    \includegraphics[width=0.98\linewidth]{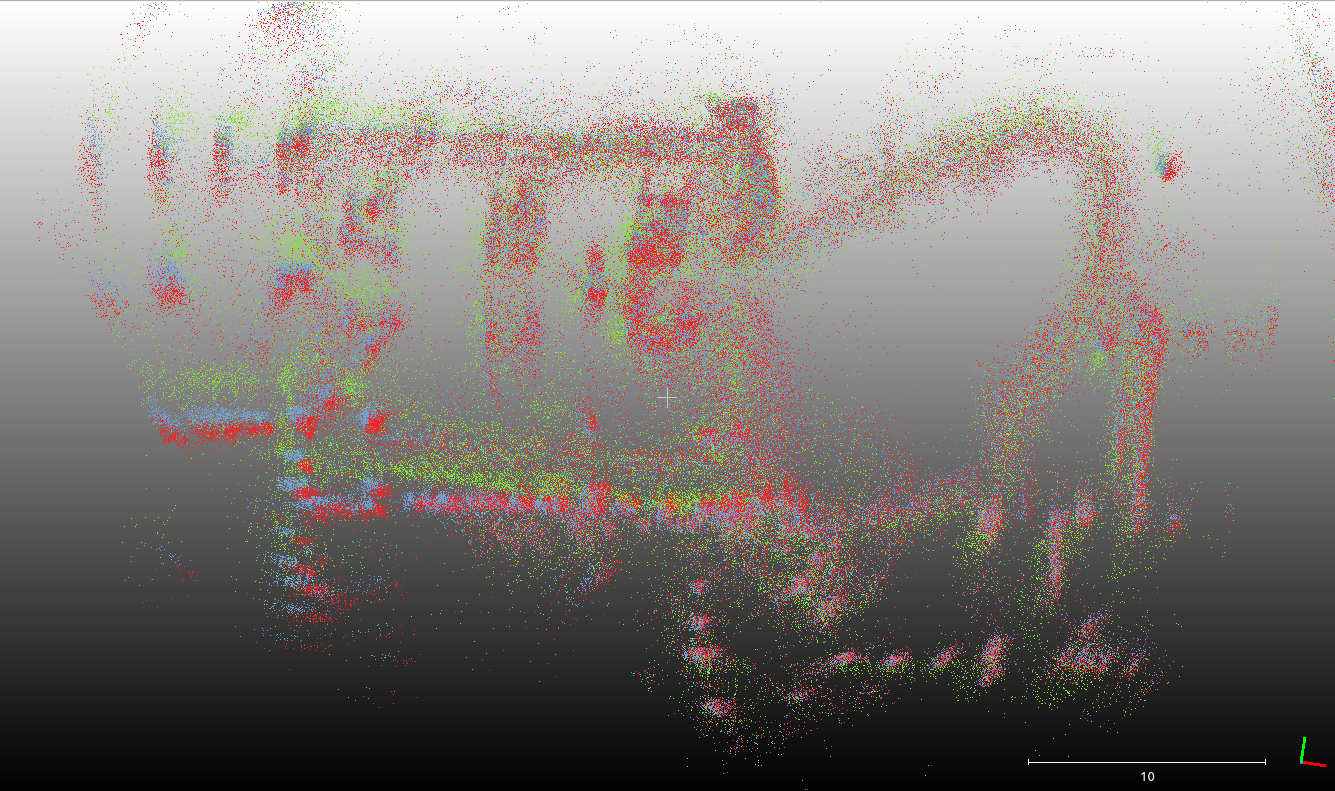}
    \includegraphics[width=0.98\linewidth]{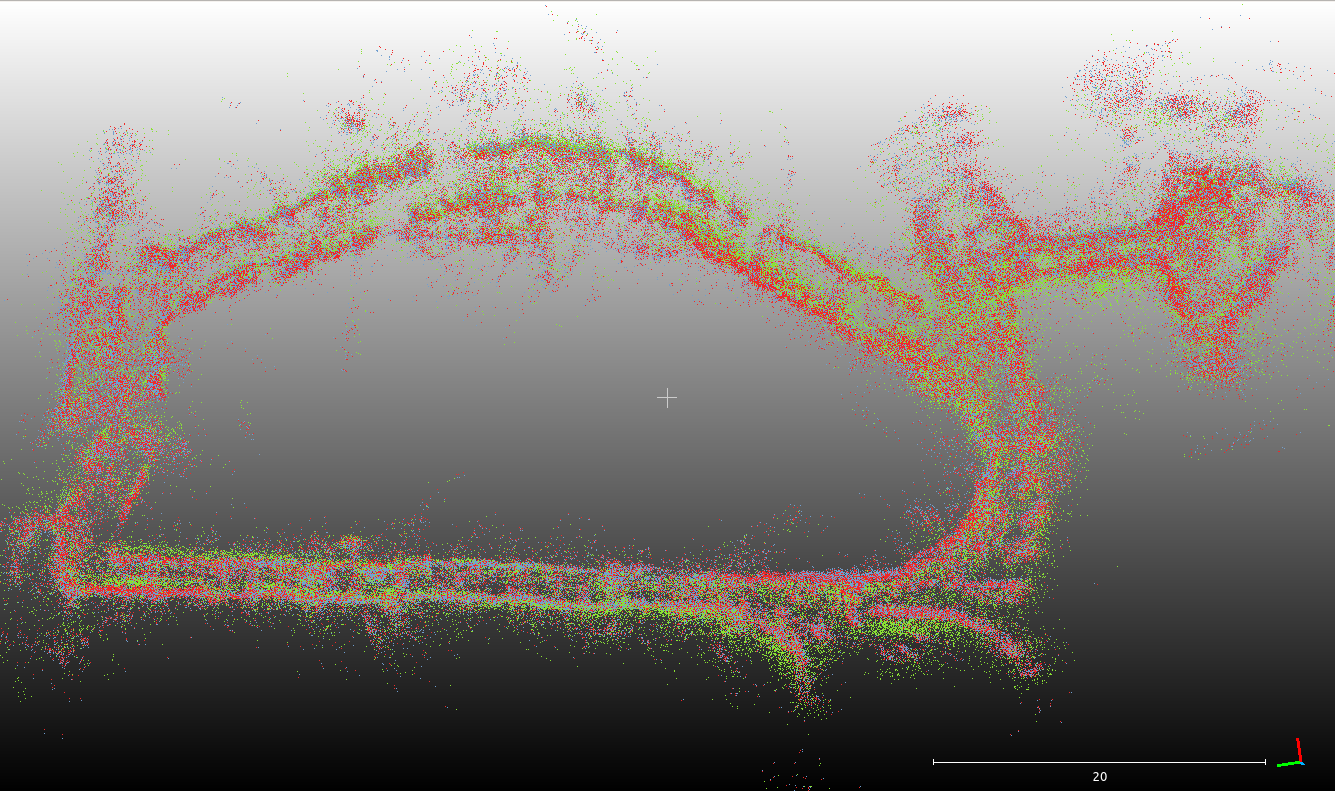}
    \caption{Mapping results on ColoRadar. Ground truth is shown in green, the map after pose graph optimization in red, and the map after six iterations of the proposed refinement in blue. Top: \textit{edgar\_classroom\_run0}. Bottom: \textit{outdoors\_run0}. The refined map is generally closer to the ground-truth geometry, especially in the left part of the top sequence.}
    \label{fig:coloradar}
\end{figure}

Figure~\ref{fig:coloradar} shows that the proposed refinement generally brings the reconstructed map closer to the ground-truth geometry than both the initial odometry and pose graph optimization. The quantitative results in Table~\ref{tab:metrics} show the same overall trend. Compared with the initial odometry, RAMBA consistently improves ATE and rotation RPE, and also improves CD-L1 on the outdoor sequence. On \textit{edgar\_classroom\_run0}, the trajectory metrics improve after optimization, while the best CD-L1 is already achieved by pose graph optimization, suggesting that this sequence is largely resolved once the main loop closure is enforced.

\begin{table}[!htbp]
\centering
\caption{Trajectory and map-quality metrics on SNAIL Radar and ColoRadar. RIO: initial radar--inertial odometry. PGO: pose graph optimization. opt2/4/6: the proposed method after 2, 4, and 6 outer iterations. Lower is better for all metrics. Metrics are reported as ATE RMSE (m), rotation RPE RMSE (deg), and Chamfer-L1 (cm). {\bf Bold} indicates the best result and \underline{underline} the second best.}
\label{tab:metrics}
\setlength{\tabcolsep}{4pt}
\begin{tabular}{l l r r r r r}
\toprule
Seq. & Metric & RIO & PGO & opt2 & opt4 & opt6 \\
\midrule
  \multirow{3}{*}{20231105/4} & ATE & 21.12 & 6.42 & 6.29 & \underline{6.11} & \textbf{5.94} \\
   & RPE & \textbf{6.24} & 7.23 & \underline{6.28} & 6.41 & 6.46 \\
   & Ch-L1 & 40.3 & \underline{39.5} & 40.0 & \textbf{39.4} & 39.8 \\
\midrule
  \multirow{3}{*}{20231105/6} & ATE & 5.30 & 1.93 & 1.50 & \underline{1.15} & \textbf{0.90} \\
   & RPE & 7.69 & 3.82 & 3.63 & \underline{3.46} & \textbf{3.29} \\
   & Ch-L1 & 38.4 & 32.5 & 30.1 & \underline{28.4} & \textbf{27.4} \\
\midrule
  \multirow{3}{*}{20231208/1} & ATE & 8.74 & 0.59 & \textbf{0.35} & \underline{0.39} & 0.46 \\
   & RPE & 7.45 & 4.25 & 1.68 & \underline{1.52} & \textbf{1.51} \\
   & Ch-L1 & 32.9 & 35.3 & \textbf{32.2} & \underline{32.2} & 32.7 \\
\midrule
  \multirow{3}{*}{edgar-classroom/0} & ATE & 2.70 & 0.51 & \textbf{0.47} & \underline{0.47} & 0.47 \\
   & RPE & 1.63 & 5.46 & \textbf{1.17} & \underline{1.17} & 1.17 \\
   & Ch-L1 & 28.8 & \textbf{20.4} & 20.7 & 20.7 & \underline{20.5} \\
\midrule
  \multirow{3}{*}{outdoors/0} & ATE & 1.23 & 0.85 & 0.79 & \underline{0.75} & \textbf{0.71} \\
   & RPE & 5.81 & 5.66 & \textbf{1.18} & 1.25 & \underline{1.23} \\
   & Ch-L1 & 25.8 & 23.2 & 22.7 & \underline{22.4} & \textbf{22.3} \\
\bottomrule
\end{tabular}
\end{table}

\subsection{Results on SNAIL Radar}\label{subsec:snail}

On SNAIL Radar \cite{huaiSNAILRadarLargescale2025}, we evaluate three sequences: 20231105/data4, 20231105/data6, and 20231208/data1. The place-recognition module finds two verified revisits on data4 and one verified revisit on each of data6 and data1.

\begin{figure*}[!htbp]
    \centering
    \includegraphics[width=0.64\linewidth]{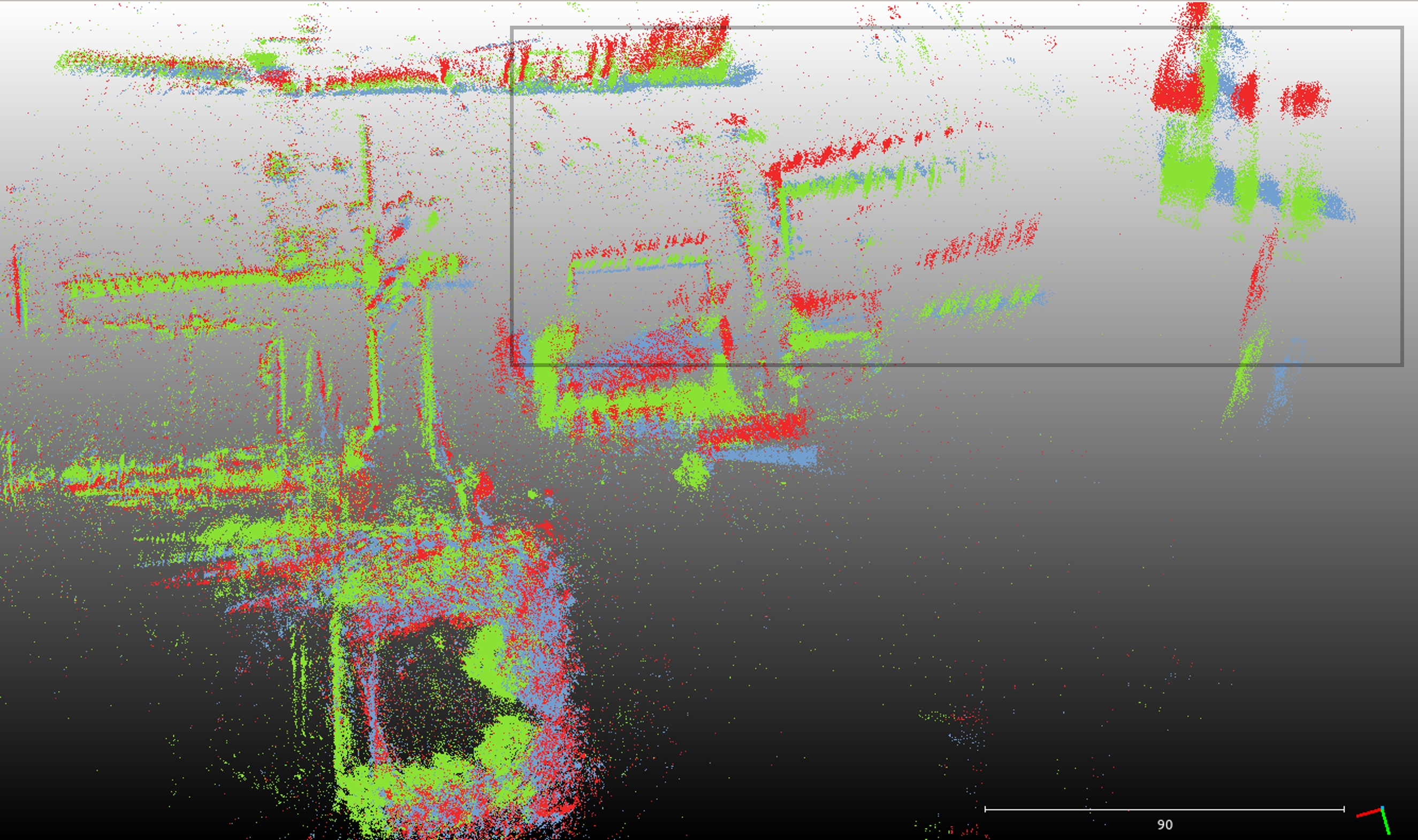}
    \includegraphics[width=0.98\linewidth]{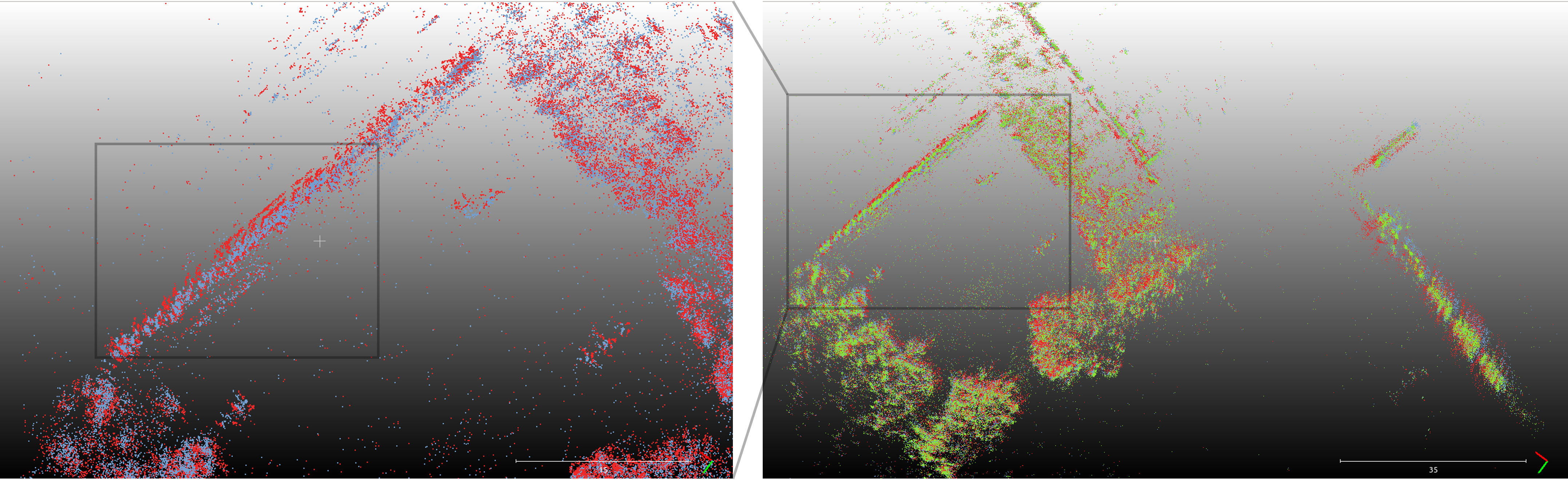}
    \includegraphics[width=0.98\linewidth]{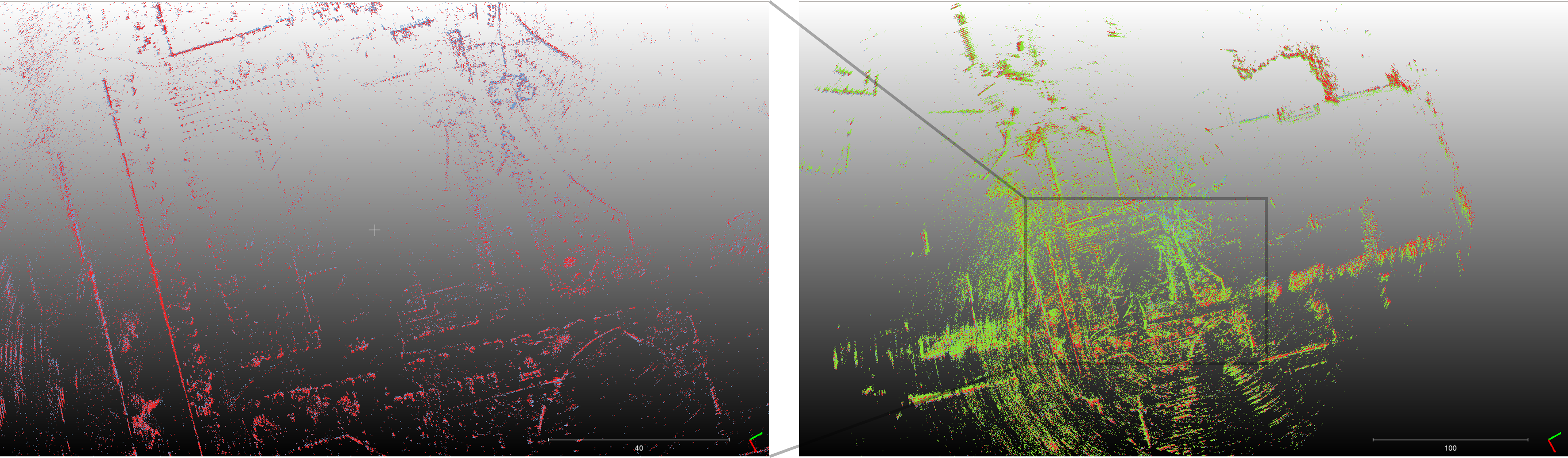}
    \caption{Mapping results on SNAIL Radar. Ground truth is shown in green, the map after pose graph optimization in red, and the map after six iterations of the proposed refinement in blue. Top: 20231105/data4. Middle: 20231105/data6. Bottom: 20231208/data1. The proposed refinement reduces structural blur and improves wall consistency, especially on the first two sequences. In the middle sequence, the wall reconstructed after pose graph optimization appears thicker than that obtained after the proposed refinement.}
    \label{fig:snail}
\end{figure*}

Figure~\ref{fig:snail} compares the maps obtained from the initial odometry, pose graph optimization, and the proposed refinement. The proposed bundle-adjustment refinement noticeably sharpens structures and reduces map blur, especially on data4 and data6. The quantitative results in Table~\ref{tab:metrics} show substantial improvements in ATE relative to the initial odometry, together with consistent gains in CD-L1 on two of the three sequences. The best overall results are typically achieved after six outer iterations, although the gains on 20231208/data1 saturate earlier.

Overall, the results on both datasets show that repeated correspondence update and bundle adjustment improve radar mapping consistency and usually also improve trajectory accuracy. In most evaluated sequences, six outer iterations provide the strongest overall performance.

\section{Conclusion}\label{sec:conclusion}

We presented RAMBA, an offline bundle-adjustment framework for consistent 4D radar mapping. The method jointly refines radar states using covariance-weighted geometric constraints, IMU preintegration, and ego-velocity measurements. The results on ColoRadar and SNAIL Radar show that the proposed refinement improves both trajectory consistency and map quality. More broadly, this study highlights the importance of generalized point-to-point constraints for mapping noisy radar point clouds, where reliable surface normals are often unavailable. This perspective may further lead to a more uniform geometric mapping framework for both LiDAR and radar point clouds.

{
	\begin{spacing}{1.17}
		\normalsize
		\bibliography{zotero} 
	\end{spacing}
}

\end{document}